\documentclass{article}

\usepackage[table]{xcolor}
\usepackage{microtype}
\usepackage{graphicx}
\usepackage{subfigure}
\usepackage{booktabs} 

\usepackage{hyperref}

\usepackage[accepted]{icml2025}

\usepackage{amsmath}
\usepackage{amssymb}
\usepackage{mathtools}
\usepackage{amsthm}

\usepackage[capitalize,noabbrev]{cleveref}

\theoremstyle{plain}

\theoremstyle{definition}

\theoremstyle{remark}

\usepackage[textsize=tiny]{todonotes}

\usepackage{xspace}
\usepackage{multirow}
\newcommand{\framework}{\textsc{NoLiMa}\xspace}
\definecolor{shadedRed}{HTML}{faa7a7}
\usepackage{float}

\usepackage{array} 
\usepackage{ragged2e} 
\usepackage{makecell} 

\icmltitlerunning{\framework: Long-Context Evaluation Beyond Literal Matching}

\newcounter{notecounter}

\newcommand{\enoteson}{\long\gdef\enote##1##2{{
\stepcounter{notecounter}
{\large\bf
\hspace{0cm}\arabic{notecounter} $<<<$ ##1: ##2
$>>>$\hspace{1cm}}}}}

\enoteson

\long\def\devour#1{}

\begin{document}

\twocolumn[
\icmltitle{\framework: Long-Context Evaluation Beyond Literal Matching}



\icmlsetsymbol{equal}{*}
\icmlsetsymbol{internship}{*}

\begin{icmlauthorlist}
\icmlauthor{Ali Modarressi}{LMU,MCML,internship}
\icmlauthor{Hanieh Deilamsalehy}{Adobe}
\icmlauthor{Franck Dernoncourt}{Adobe}
\icmlauthor{Trung Bui}{Adobe}
\icmlauthor{Ryan Rossi}{Adobe}
\icmlauthor{Seunghyun Yoon}{Adobe}
\icmlauthor{Hinrich Schütze}{LMU,MCML}

\end{icmlauthorlist}

\icmlaffiliation{LMU}{Center for Information and Language Processing, LMU Munich, Germany}
\icmlaffiliation{MCML}{Munich Center for Machine Learning (MCML)}
\icmlaffiliation{Adobe}{Adobe Research}

\icmlcorrespondingauthor{Ali Modarressi}{amodaresi@cis.lmu.de}

\icmlkeywords{Machine Learning, ICML}

\vskip 0.3in
]



\printAffiliationsAndNotice{\textsuperscript{*}Work done during an internship at Adobe Research.}

\begin{abstract}
Recent large language models (LLMs) support long contexts ranging from 128K to 1M tokens. A popular method for evaluating these capabilities is the needle-in-a-haystack (NIAH) test, which involves retrieving a ``needle'' (relevant information) from a ``haystack'' (long irrelevant context). Extensions of this approach include increasing distractors, fact chaining, and in-context reasoning. However, in these benchmarks, models can exploit existing literal matches between the needle and haystack to simplify the task. 
To address this, we introduce \framework, a benchmark extending NIAH with a carefully designed needle set, where questions and needles have minimal lexical overlap, requiring models to infer latent associations to locate the needle within the haystack. We evaluate 13 popular LLMs that claim to support contexts of at least 128K tokens. While they perform well in short contexts ($<$1K), performance degrades significantly as context length increases. At 32K, for instance, 11 models drop below 50\% of their strong short-length baselines. Even GPT-4o, one of the top-performing exceptions, experiences a reduction from an almost-perfect baseline of 99.3\% to 69.7\%. Our analysis suggests these declines stem from the increased difficulty the attention mechanism faces in longer contexts when literal matches are absent, making it harder to retrieve relevant information. 
Even models enhanced with reasoning capabilities or CoT prompting struggle to maintain performance in long contexts.
We publicly release the dataset and evaluation code at \href{https://github.com/adobe-research/NoLiMa}{https://github.com/adobe-research/NoLiMa}.\footnote{\label{update_footnote}Recent models not covered in the main body of this paper (such as GPT-4.1) have
improved long-context performance. However, the main finding of this paper also
holds for these newer models: performance on \framework starts declining rapidly
on contexts that are relatively short compared to their claimed context length. See Appendix~\ref{sec:appx_beyond_32K} for detailed results.}
\end{abstract}

\section{Introduction}
In recent years, large language models (LLMs) have made remarkable advancements in handling long-context inputs \cite{chen2023extending, xiong-etal-2024-effective, peng2024yarn}. This capability has unlocked new possibilities in various NLP tasks that require understanding or generating content over extended documents. Examples include long- or multi-document question answering (QA), summarization, and many-shot in-context learning \cite{lee2024longcontextlanguagemodelssubsume, chang2024booookscore, agarwal2024manyshot}. To evaluate these models' effectiveness in handling long contexts, several benchmarks have been developed.
One prominent benchmark is Needle-in-a-Haystack (NIAH),
which tests a model's ability to search for and retrieve a
specific fact (the ``needle'') hidden within irrelevant
information (the ``haystack'') \cite{kamradt2023needle, mohtashami2023randomaccess}. While the baseline NIAH task assesses surface-level retrieval capabilities, recent adaptations have increased its complexity. These enhancements include introducing multiple needles, incorporating additional distractor material, and interconnecting facts to necessitate in-context reasoning (e.g., fact-chaining) \cite{hsieh2024ruler, levy-etal-2024-task, kuratov2024babilong}. 
Other benchmarks, such as long-, multi-document QA, and long conversation understanding, have also been proposed to evaluate long-context comprehension in a more downstream task manner \cite{liu-etal-2024-lost, yen2024helmet, zhang2024inftybenchextendinglongcontext, dong-etal-2024-bamboo, wang-etal-2024-leave, maharana-etal-2024-evaluating}.

Arguably, these tasks share a common foundation: the ability to recall previously seen information \cite{goldman-etal-2024-really}. This broader category, termed association recall tasks, has been extensively studied in machine learning \cite{graves2014neuralturingmachines, ba2016using}. A key argument is that the attention mechanism, which is the underlying foundation of many LLMs, is inherently adept at identifying and recalling associations present in the input \cite{olsson2022incontextlearninginductionheads, arora2024zoology}. However, this raises an important question: Long-context benchmarks feature tasks where the queried input (e.g., a question or a task) has literal matches with the provided context. \textit{Do such literal
matches make it easier for language models to locate relevant information and
output correct answers?}

We argue that many existing long-context benchmarks either explicitly (e.g., synthetic tasks or NIAH-based) or implicitly (e.g., multi-document or long-document QA) contain such literal matches. To address this, we introduce \textbf{\framework}, a benchmark designed to minimize literal overlap between questions and their corresponding needles. In \framework, questions and needles contain keywords that are related through associative links, such as real-world knowledge or commonsense facts. By embedding these needles in a haystack, \framework challenges models to leverage latent associative reasoning capabilities rather than relying on surface-level matching.

We evaluate \framework over 13 state-of-the-art language models, all claiming to support token lengths of at least 128K, including GPT-4o, Gemini 1.5 Pro, and Llama 3.3 70B \cite{hurst2024gpt, team2024gemini, meta2024llama3_3}. Unlike NIAH-based evaluations, which contain literal matches and exhibit near-saturated performance, \framework presents a more demanding challenge that highlights the limitations of these models.
Their performance declines noticeably as context length increases, with considerable drops even at 2K–8K tokens. For instance, at 32K tokens, 11 out of 13 models achieve only half of their short-context performance.

We conduct extensive analyses using \framework, yielding the following insights:
\begin{itemize}
    \item \textbf{Impact of Latent Hops and Fact Direction}: We demonstrate how the number of associative reasoning steps (latent hops) and the ordering of elements within a fact statement influence task performance. (\S~\ref{sec:latent_hops_&_inversion})
    \item \textbf{Context Length vs. Needle Position}: Our aligned-depth analysis shows that as latent reasoning complexity grows, performance depends more on context length than needle position. Without surface cues, longer contexts overwhelm the attention mechanism. (\S~\ref{sec:depth_analysis})
    \item \textbf{Chain-of-Thought (CoT) Prompting and Reasoning-based Models}: While CoT prompting or reasoning-based models such as GPT-o1 \cite{jaech2024openai} improve performance by encouraging step-by-step reasoning, they fail to fully mitigate the challenge, particularly in contexts exceeding 16K tokens. (\S~\ref{sec:cot_prompting})
    \item \textbf{Ablation Tests}: We confirm that the presence of literal matches significantly simplifies the task, enabling models to achieve high accuracy in answering questions. In contrast, when literal matches serve as distractors, they severely impair accuracy. (\S~\ref{sec:literal_match_effect})
\end{itemize}
Through \framework, we reveal the limitation of literal matching in long-context benchmarks and introduce a novel approach for evaluating models' latent reasoning in longer contexts.

\section{Related Work}

\begin{table}[t!]
	\setlength\tabcolsep{3.8pt}
        \small
	\centering
	\begin{tabular}{lccc}
		\toprule
          & {\textbf{R-1}} & {\textbf{R-2}} & {\textbf{R-L}} \\
        \midrule
        \multicolumn{4}{c}{\emph{Long-document QA}} \\
        $\infty$Bench QA \cite{zhang2024inftybenchextendinglongcontext} & 0.966 & 0.545 & 0.960 \\
        $\infty$Bench MC \cite{zhang2024inftybenchextendinglongcontext} & 0.946 & 0.506 & 0.932 \\
        \midrule
        \multicolumn{4}{c}{\emph{RAG-style (Multi-doc) QA}} \\
        RULER QA \cite{hsieh2024ruler} & 0.809 & 0.437 & 0.693 \\ 
        HELMET (RAG) \cite{yen2024helmet} & 0.689 & 0.304 & 0.555 \\
        \midrule
        \multicolumn{4}{c}{\emph{Recall-based}} \\
        Vanilla NIAH \cite{kamradt2023needle} & 0.905 & 0.789 & 0.855 \\
        RULER S-NIAH \cite{hsieh2024ruler} & 0.571 & 0.461 & 0.500 \\
        BABILong (0K) \cite{kuratov2024babilong} & 0.553 & 0.238 & 0.522 \\
        \midrule
        \textbf{\framework} & \textbf{0.069} & \textbf{0.002} & \textbf{0.067} \\
		\bottomrule
	\end{tabular}
	\caption{ROUGE precision scores between the input document and the question: higher ROUGE scores indicate greater literal matches between the question and the relevant context.}
	\label{tab:literal_match_rouge}
\end{table}
With the increasing popularity of long-context language modeling, numerous benchmarks have been introduced to evaluate this capability. Needle-in-a-Haystack (NIAH) is the most well-known and widely used benchmark \cite{mohtashami2023randomaccess, kamradt2023needle}. However, due to performance saturation, various extensions have been proposed. These include increasing complexity by adding more needles, chaining needles to require inter-needle reasoning (fact-chaining), or incorporating arithmetic or code reasoning \cite{kamradt2023needle, hsieh2024ruler, levy-etal-2024-task, kuratov2024babilong, hengle2024multilingual, zhang2024inftybenchextendinglongcontext, vodrahalli2024michelangelolongcontextevaluations}.
Some tasks increase the complexity to such an extent that
they become overly difficult even in short-context
scenarios. For instance, BABILong includes tasks that
perform poorly (e.g., the counting task achieves 28\% accuracy) even without any irrelevant background text (0K) \cite{kuratov2024babilong}. Similarly, the Ancestral Tree Challenge (ATC) employs extensive fact-chaining, resulting in tasks that are overly complex even for short contexts ($<$1K) \cite{li2024needlebenchllmsretrievalreasoning}.
While such tasks challenge language models in long contexts, they raise the question of whether the tasks are inherently too complex for models to handle, regardless of context length.

\devour{
\begin{table*}[t]
	\small
	\setlength\tabcolsep{5pt}
	\centering
    \begin{tabular}{p{0.25\linewidth} p{0.05\linewidth} p{0.37\linewidth} p{0.02\linewidth} p{0.18\linewidth}}
        \toprule
        \textbf{Question} & \multicolumn{2}{l}{\textbf{Needles}} & \multicolumn{2}{l}{\textbf{Keyword Types}}\\
        \midrule
        \multirow{2}{*}{Which character has been to $W_q$?} & Def. & Actually, [CHAR] lives next to the $W_n$. & $W_n$ & Buildings \& Landmarks\\
         & Inv. & $W_n$ is next to where [CHAR] lives. & $W_q$ & Countries, cities, states \\
        \bottomrule
    \end{tabular}
	\caption{An example template of the proposed needle set in \framework (all templates are available in Appendix \ref{sec:appx_needle_set}.) The placeholders [CHAR], $W_q$, and $W_n$ represent the randomly selected character (also the answer), the query keyword, and the needle keyword, respectively. Def.:
      default order. Inv.: inverted order.}
	\label{tab:needle_set_one}
\end{table*}
}

\def\mrspace{0.1cm}
\def\bigspace{0.2cm}

\begin{table*}[t]
	\small
	\centering
    \begin{tabular}{l@{\hspace{\bigspace}}l@{\hspace{\mrspace}}l@{\hspace{\bigspace}}l@{\hspace{\mrspace}}l}
        \toprule
        \textbf{Question} & \multicolumn{2}{l}{\textbf{Needle}} & \multicolumn{2}{l}{\textbf{Keyword Type}}\\
        \midrule
        \multirow{2}{*}{Which character has been to $W_q$?}
        & (Def.) & Actually, [CHAR] lives next to the $W_n$. &
        $W_n$ & Buildings\&Landmarks (e.g., Semper Opera)\\
         & (Inv.) & $W_n$ is next to where [CHAR] lives. &
        $W_q$ & Countries, cities, states (e.g., Dresden) \\
        \bottomrule
    \end{tabular}
	\caption{An example template of the proposed needle set in \framework. (All templates are available in Appendix \ref{sec:appx_needle_set}.) The placeholders [CHAR], $W_q$, and $W_n$ represent the randomly selected character (also the answer), the query keyword, and the needle keyword, respectively. Def.:
      default order. Inv.: inverted order.}
	\label{tab:needle_set_one}
\end{table*}

\paragraph{Literal Matching in Long-Context Benchmarks.}
\label{sec:related_literal_matching}
Another frequent pattern in many long-context benchmarks is the presence of literal matches between the facts required to answer a question and the question itself. This fact is not limited to synthetic recall-based tasks (e.g., vanilla NIAH, RULER retrieval-based sets) but also affects downstream-like QA-based benchmarks \cite{hsieh2024ruler, liu-etal-2024-lost, zhang2024inftybenchextendinglongcontext, bai-etal-2024-longbench, yen2024helmet}, which often implicitly include literal matches between the relevant document and the question. Although many of these studies introduce complexity by adding similar documents as distractors, literal matches can still provide cues. These cues may help models focus on potential relevant facts based on matches, as attention mechanisms excel at recalling repetitive patterns \cite{olsson2022incontextlearninginductionheads, arora2024zoology}.
We later demonstrate to what extent literal matches simplify
recall-based questions
(cf. \ref{sec:literal_match_effect}). To quantify the
prevalence of these matches in popular benchmarks, we
compute ROUGE (R-1, R-2, and R-L) precision
scores\footnote{We use precision as our metric to measure
how many of the question's tokens occur in the relevant
context, rather than the reverse.} \citep{lin-2004-rouge}
between the question and the context -- the needle (in recall-based tasks), the relevant document (in multi-document setups), or the full document (in long-document QA). This analysis measures the degree of literal overlap between the question and the context.
Table \ref{tab:literal_match_rouge} demonstrates
that \framework has
much less literal overlap than other datasets.

\section{\framework}
\label{sec:NOLIMA}
The goal of \framework is to design a task that is
inherently simple to solve through associative reasoning,
but for which surface-level matching has zero utility. As a
result, \framework allows us to cleanly investigate
associative reasoning in long-context scenarios without
confounding from surface-level effects.

The main elements of \framework are similar to vanilla NIAH.
A ``needle'' -- a single key piece of information --
is placed within a ``haystack'', i.e., a long irrelevant
text (in our case, snippets from books). Given a
question, the model is then tested on its ability to find
the needle.    The needle is designed to be a clearly relevant
answer to the question.  In contrast to existing NIAH tasks, we
impose the condition that the question have minimal literal match
with the needle. To achieve this, we design a set of needles 
and corresponding questions, collectively referred to as
a ``needle set.''  Table \ref{tab:needle_set_one} presents
one of the constructed needle set templates (see Appendix
\ref{sec:appx_needle_set} for the full list).  Each needle
consists of a unique character and specific information
about them. Example:
\begin{center}
Actually, Yuki lives next to the \textbf{Semper Opera House}. 
\end{center}
The needle 
contains a keyword ($W_n$, here ``Semper Opera House'') that serves as the critical link
between needle and question. The question is
designed to retrieve this information by asking which
character possesses a specific attribute $W_q$, ``Dresden''
in the example:
\begin{center}
Which character has been to \textbf{Dresden}?
\end{center}
The Semper Opera House is located in Dresden.
Thus, the
model should 
be able to identify the latent association link between
$W_q$ (``Dresden'') in the question
and $W_n$ (``Semper Opera House'')
in the needle.
Since there is no literal overlap between
needle and question, the model
must rely on this latent association link to
retrieve ``Yuki'', the correct answer.
For some of our needles,
the association involves commonsense
reasoning instead of world knowledge. Example: ``Then Yuki
mentioned that he has been vegan for years.'' $\rightarrow$
``Which character cannot eat fish-based meals?''  To push the
limits of the model’s ability to identify hidden
associations, we include questions that require
two hops to connect $W_q$ with $W_n$, for example:
\begin{center}
Which character has been to \textbf{the state of Saxony}?
\end{center}
Here, the model should tap into its knowledge that
Dresden (and hence the Semper Opera) is located
in the state of Saxony.
This two-hop setup further increases the difficulty of identifying
the latent association of $W_q$ with $W_n$.

To make \framework an effective benchmark for evaluating LLM
long-context abilities, we impose several constraints on
the needle set.  (i) We select keywords that
ensure simplicity -- so that, without irrelevant context,
the associations are clear and the model can identify the
correct answer.  (ii) We randomize the assignment of
character names from a diverse pool to minimize sensitivity
to tokenization problems and mitigate ethnic bias \cite{biasesinLLMs2023Navigli, jiang-etal-2024-peek}. Names
already occurring in the haystacks are excluded.  (iii) We ensure $W_n$ is uniquely associated with $W_q$, avoid language-based cues, and in most cases employ preface phrases--short lead-ins or contextual buildup (e.g., ``Actually,'' ``In 2013, after waiting in line...'')--to isolate needles from preceding context.
See Appendix
\ref{sec:appx_needle_set} for details.
\begin{figure*}[t!]
    \centering
    {\includegraphics[width=0.92\linewidth, trim=1 1 1 1, clip] {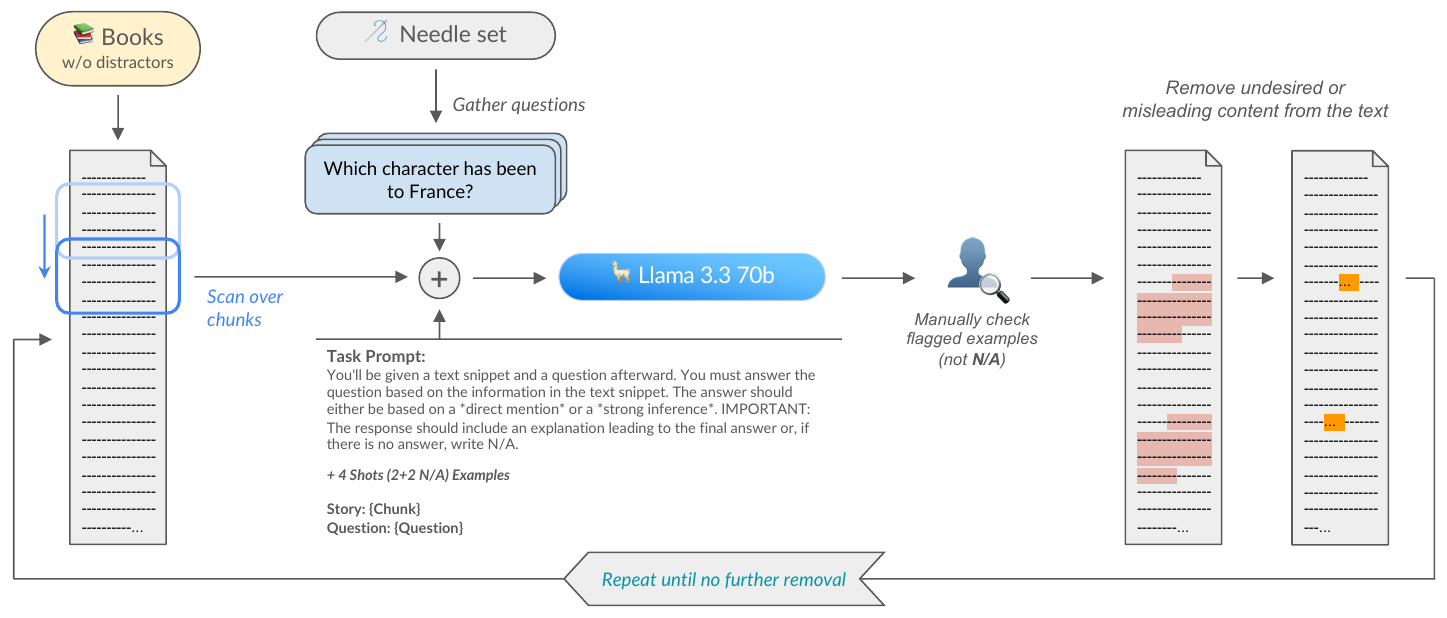}}
    \caption{Haystack filtering pipeline for undesired or misleading content}
    \label{fig:haystack_pipeline}
\end{figure*}

\subsection{Haystack Filtering Pipeline}
\label{sec:haystack_filtering}
We devise a filtering process
to ensure that the haystack does not contain: (1) Any
distracting words that have extreme literal or high semantic
similarity with the key points mentioned in the question,
(2) Any information that explicitly or through inference is a potential false answer to the question.

\paragraph{Distractor Filtering.} For this step, we use an embedding function, Contriever \citep{izacard2022unsupervised}, to find similar words in the haystack to the keywords of the questions. First we gather all words in the haystack and compute their respective embedding. Then using dot-product similarity we compute their similarity to the question keywords. 
We manually inspect the top-20 similar words per each $W_q$ and flag those with high semantic or substring similarity for removal.
In the removal process those sentences that contain flagged words are removed from the haystack. 
This initial filtering step helps to avoid an uncontrolled set of superficial distractors that could undesirably disrupt the experimental results. We will discuss the impact of distractors on the model performance in our analysis (Section \ref{sec:literal_match_effect}).

\paragraph{Filtering Undesired Answer Candidates.} In this step, we implement a semi-automatic redaction process to detect and remove text spans that could be interpreted as plausible but unintended answers. As shown in Figure \ref{fig:haystack_pipeline}, this process takes the haystack text—already filtered for distractors—along with questions from our needle set as input.
Assuming the model should infer cases within short contexts, we scan the input texts in smaller chunks.\footnote{With an 800-character stride and a 1000-character chunk size ($\sim$250 tokens).} To identify potential answers within a chunk, we pair each question with the chunk and input them into an instruction-tuned language model, along with a short instruction and few-shot examples. The model responds with either ``N/A'' (indicating no relevant information was found) or an explanation identifying a possible candidate answer. Flagged examples are manually reviewed\footnote{All manual reviews---in both filtering steps---were conducted by one of the authors.} to determine whether the identified content should be removed. If no relevant content is identified, the text remains unchanged. This process is repeated across all selected haystacks until no further removals are necessary.

\section{Experiments}

\begin{table*}[ht!]
        \small
	\setlength\tabcolsep{5pt}
	\centering
	\begin{tabular}{l|cc|ccccccc}
		\toprule
        \multirow{2}{*}{\textbf{Models}} & \textbf{Claimed} & \textbf{Effective} & \textbf{Base Score} & \multirow{2}{*}{\textbf{1K}} & \multirow{2}{*}{\textbf{2K}} & \multirow{2}{*}{\textbf{4K}} & \multirow{2}{*}{\textbf{8K}} & \multirow{2}{*}{\textbf{16K}} & \multirow{2}{*}{\textbf{32K}} \\
        & \textbf{Length} & \textbf{Length} & ($\times$0.85: \textbf{Thr.}) &&&&&& \\
        \midrule
        GPT-4o              & 128K & 8K & 99.3 (84.4) & \underline{98.1} & \underline{98.0} & \underline{95.7} & \underline{89.2} & 81.6 & 69.7 \\
        Llama 3.3 70B       & 128K & 2K & 97.3 (82.7) & \underline{94.2} & \underline{87.4} & 81.5 & 72.1 & 59.5 & \cellcolor[HTML]{faa7a7}42.7 \\
        Llama 3.1 405B      & 128K & 2K & 94.7 (80.5) & \underline{89.0} & \underline{85.0} & 74.5 & 60.1 & 48.4 & \cellcolor[HTML]{faa7a7}38.0 \\
        Llama 3.1 70B       & 128K & 2K & 94.5 (80.3) & \underline{91.0} & \underline{81.8} & 71.2 & 62.7 & 51.8 & \cellcolor[HTML]{faa7a7}43.2 \\
        Gemini 1.5 Pro      & 2M   & 2K & 92.6 (78.7) & \underline{86.4} & \underline{82.7} & 75.4 & 63.9 & 55.5 & 48.2 \\
        Jamba 1.5 Mini      & 256K & $<$1K & 92.4 (78.6) & 76.3 & 74.1 & 70.8 & 62.2 & 52.7 & \cellcolor[HTML]{faa7a7}43.6 \\
        Command R+          & 128K & $<$1K & 90.9 (77.3) & 77.0 & 73.5 & 66.2 & \cellcolor[HTML]{faa7a7}39.5 & \cellcolor[HTML]{faa7a7}21.3 & \cellcolor[HTML]{faa7a7}7.4 \\
        Gemini 2.0 Flash    & 1M   & 4K & 89.4 (76.0) & \underline{87.7} & \underline{87.5} & \underline{77.9} & 64.7 & 48.2 & \cellcolor[HTML]{faa7a7}41.0 \\
        Mistral Large 2     & 128K & 2K & 87.9 (74.7) & \underline{86.1} & \underline{85.5} & 73.3 & 51.4 & \cellcolor[HTML]{faa7a7}32.6 & \cellcolor[HTML]{faa7a7}18.8 \\
        Claude 3.5 Sonnet   & 200K & 4K & 87.5 (74.4) & \underline{85.4} & \underline{84.0} & \underline{77.6} & 61.7 & 45.7 & \cellcolor[HTML]{faa7a7}29.8 \\
        Gemini 1.5 Flash    & 1M   & $<$1K & 84.7 (72.0) & 68.6 & 61.6 & 51.0 & 44.4 & \cellcolor[HTML]{faa7a7}35.5 & \cellcolor[HTML]{faa7a7}28.6 \\
        GPT-4o mini         & 128K & $<$1K & 84.8 (72.1) & 67.7 & 58.2 & 44.2 & \cellcolor[HTML]{faa7a7}32.6 & \cellcolor[HTML]{faa7a7}20.6 & \cellcolor[HTML]{faa7a7}13.7 \\
        Llama 3.1 8B        & 128K & 1K  & 76.7 (65.2) & \underline{65.7} & 54.4 & 44.1 & \cellcolor[HTML]{faa7a7}31.9 & \cellcolor[HTML]{faa7a7}22.6 & \cellcolor[HTML]{faa7a7}14.2 \\
		\bottomrule
	\end{tabular}
	\caption{\framework benchmark results on the selected models. Following \citet{hsieh2024ruler}, we report the effective length alongside the claimed supported context length for each model. However, we define the effective length as the maximum length at which the score remains above a threshold, set at 85\% of the model's base score (shown in parentheses). Scores exceeding this threshold are \underline{underlined}. Scores that are below 50\% of the base score are shaded in \colorbox{shadedRed}{red}.}
	\label{tab:models_performance}
\end{table*}

\subsection{Dataset Configuration}
\label{sec:dataset_config}
In \framework, we use 5 groups of needles,
each with two 
``word order'' variations: 
\textit{default} and \textit{inverted}. In the default order, the 
answer character (CHAR) precedes the needle keyword ($W_n$), following the pattern ``\ldots [CHAR] \ldots $W_n$'' (see Table~\ref{tab:needle_set_one}, column ``Needle'').
In the inverted order, the character name follows $W_n$, yielding the pattern ``$W_n$ \ldots [CHAR] \ldots''.
Each group includes 2–6 keyword sets, with some sets containing multiple $W_q$ items to produce both one-hop and two-hop examples. This setup results in 58 question-needle pairs in total.
To generate the haystacks, we select 10 open-license books, ensuring each covers at least 50K tokens. Using the filtering mechanism described in Section \ref{sec:haystack_filtering}, we process the text to prepare it for haystack construction. To mitigate potential memorization issues—since these books are publicly available—we construct haystacks by concatenating short snippets. Specifically, we iteratively and randomly select a book, extract a continuous snippet (under 250 tokens), and append it to the haystack until it exceeds 2K lines, resulting in haystacks exceeding 60K tokens.
In all experiments, each needle is placed 26 times at equal intervals across the evaluated context length. With 5 randomly generated haystacks, 58 question-needle pairs, and 26 placements per context length, this setup results in 7,540 tests per context length experiment.

\subsection{Models}
For the filtering process, we opted for using the Llama 3.3 70b instruction tuned model \citep{meta2024llama3_3}. As a control test, for each question, we place its needle in 100 randomly selected chunks to determine whether the model (1) understands the filtering task and (2) is familiar with the facts and capable of inferring the answer. The model achieves a score of 99.8\% in this test, indicating its ability to effectively flag conflicting information from the haystacks.

For the evaluation process, we select five closed-source
models: GPT-4o, GPT-4o Mini \cite{hurst2024gpt},
Gemini 1.5 Pro Flash, Gemini 2.0 Flash \cite{team2023gemini, team2024gemini} and Claude 3.5 Sonnet \cite{anthropic2024claude}, along with seven open-weight Llama models: The Llama 3.x model family (3.1 8B, 70B, 405B, and 3.3 70B) \cite{dubey2024llama, meta2024llama3_3}, Mistral Large \cite{mistralLarge2411}, Command R+ \cite{cohere_for_ai_2024}, and Jamba 1.5 Mini \cite{team2024jamba}. All these models are well-known and widely used in long-context setups. 
In our analysis on reasoning-based prompting and models, we evaluate GPT-o1, GPT-o3 Mini \cite{jaech2024openai, o3Mini2024SystemCard}, and DeepSeek-R1 Distill-Llama-70B \cite{deepseekai2025deepseekr1incentivizingreasoningcapability}.
More details regarding model versions and deployment details are described in Appendix \ref{sec:appx_models}.

\subsection{Evaluation Setup \& Metric}
During inference, we use a task template (see Appendix \ref{sec:appx_task_prompts}) that instructs the model to answer the question based on the provided text. Since all questions seek the name of the character mentioned in the needle, any returned answer containing the correct name is considered accurate. Accuracy is reported as the proportion of tests with correct answers.

Models are evaluated on all tasks over context lengths of 250, 500, 1K, 2K, 4K, 8K, 16K, and 32K.
To take into account how models would perform on \framework regardless of
long-context scenario, we control the difficulty of the task by reporting a \textbf{base score}. Evaluations at context lengths of 250, 500, and 1K are used to compute the base score. These three are the shortest contexts. If a model can solve the task at these lengths, then any deterioration of its performance at greater lengths is expected to be solely due to its difficulties with generalizing over long contexts.
For each question-needle example, we compute the average
scores over 5 haystacks, then take its maximum score across the 250, 500, and 1K tests. The final base score is obtained by averaging these maximum scores across all question-needle examples.
Inspired by \citet{hsieh2024ruler}, we also report the
\textbf{effective length} of each model. While they use the
performance of Llama 2  at 4K context length as a threshold (85.6\%), we define the threshold as 85\% of the base score. Thus, the effective length of a model is the largest tested length that exceeds this threshold.
Additionally, some plots show the \textbf{normalized score}, calculated by dividing the accuracy score by the base score.

\subsection{Results}
Table \ref{tab:models_performance} presents the performance results of all \framework tests on the selected models. Most models achieve high base scores, indicating that the designed needle set is relatively simple to answer in shorter contexts.
Even models with base scores exceeding 90.0\% exhibit a significantly shorter effective length than their claimed lengths, generally limited to $\leq$2K tokens, with GPT-4o being an exception. While GPT-4o demonstrates strong overall performance, it fails to generalize effectively beyond 8K tokens.\textsuperscript{\ref{update_footnote}} Out of the 13 models, 11 exhibit performance at 32K lengths that is half or less of their base scores. For comparison, in other benchmarks with similar settings, such as BABILong (QA1) \cite{kuratov2024babilong} and RULER \cite{hsieh2024ruler}, Llama 3.1 70B achieves effective lengths of 16K\footnote{In BABILong, the effective length is also based on 85\% of the 0K base performance threshold} and 32K, respectively. However, in \framework, Llama 3.1 70B has an effective length of only 2K and shows a significant drop in performance at 32K lengths (42.7\% vs. 94.3\% base score).
Models such as Claude 3.5 Sonnet, Gemini 1.5 Flash, GPT-4o mini, and Llama 3.1 8B may have weaker base scores, but their effective lengths are calculated relative to these scores. This reveals an interesting observation: Models like Claude 3.5 Sonnet and Gemini 2.0 Flash, despite having a lower base score, may underperform in shorter contexts but demonstrate better length generalization than models with higher base scores, such as Llama 3.1 70B and Llama 3.3 70B. In fact, both Sonnet and Gemini 2.0 Flash  achieve even higher raw scores in 4K-token experiments compared to some higher-base-score models.

Model scaling generally improves performance, as seen in the progression from Llama 3.1 8B to 70B, Gemini 1.5 Flash to Pro, or GPT-4o mini to GPT-4o. However, the benefits of scaling diminish at larger scales; for example, the performance gap between Llama 3.1 70B and 405B is smaller (and sometimes worse) than that between 8B and 70B. In general, ``lite'' models such as Gemini 1.5 Flash, GPT-4o mini, and Llama 3.1 8B perform well in shorter contexts ($<$1K tokens) but fail to generalize effectively in longer contexts.

Based on the dataset construction method described in Section~\ref{sec:dataset_config}, \framework can generate haystacks at any desired length. We applied this to test GPT-4o and Gemini 2.0 Flash at 64K and 128K tokens. GPT-4o maintained over 50\% of its base score at 128K, while Gemini 2.0 Flash dropped to just 16.4\%. Full results and evaluation details are provided in Appendix~\ref{sec:appx_beyond_32K}.

\begin{figure}[t!]
    \centering
    \subfigure[One-hop vs. Two-hop]{
        \centering
        \includegraphics[width=0.46\linewidth, trim=10 9 10 8, clip] {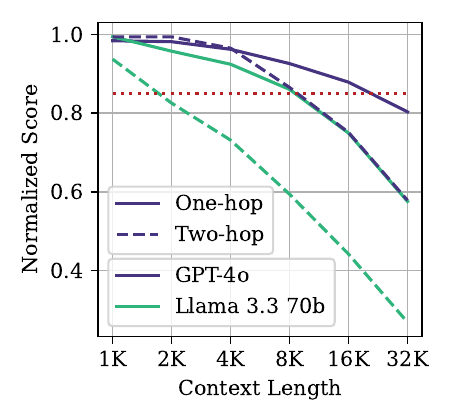}
        \label{fig:one_vs_twohop}
    }
    \hfill
    \subfigure[Default vs. Inverted]{
        \centering
        \includegraphics[width=0.46\linewidth, trim=10 9 10 8, clip] {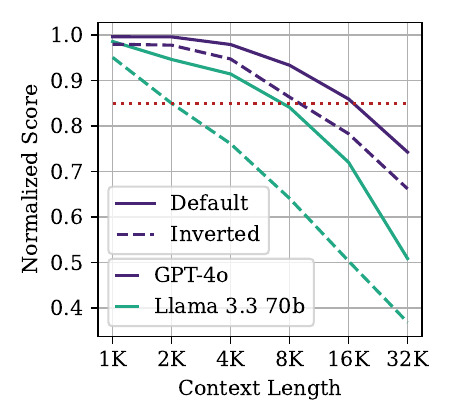}
        \label{fig:normal_vs_inverted}
    }
    \caption{Impact of (a) number of hops and (b) inversion
      of order (``[CHAR] \ldots
$W_n$'' vs.\ 
``$W_n$ \ldots CHAR'')
      on normalized performance across GPT-4o and Llama 3.3 70B models. The red dotted line indicates the 0.85 effective threshold.}
    \label{fig:hops_and_inversion}
\end{figure}

\subsubsection{Latent hops \& Inversion}
\label{sec:latent_hops_&_inversion}
As discussed in Section \ref{sec:NOLIMA}, our needle set also includes examples requiring two-hop associative linking from the question keyword to the needle keyword. To evaluate the impact on length generalization, Figure \ref{fig:one_vs_twohop} presents the normalized performance of two top-performing models on one-hop and two-hop tasks. It is evident that, for the same context lengths, questions involving two-hop latent reasoning steps are more challenging than those requiring one-hop reasoning. Notably, the performance gap between one-hop and two-hop tasks widens with increasing context lengths. GPT-4o demonstrates impressive generalization performance, handling both types of examples effectively even at context lengths up to 4K. A detailed breakdown of performance on one-hop and two-hop examples is provided in Appendix~\ref{sec:appx_one_and_two_hop_performance}, complementing the aggregate results shown in Table~\ref{tab:models_performance}.

Each group of needles includes both a default and an inverted template and Figure~\ref{fig:normal_vs_inverted} shows that inverted examples are more challenging to answer. We argue this difficulty arises from the model’s causal attention mechanism, particularly in longer contexts where attention signals weaken. 
In the default template, the question -- in particular $W_q$
-- can link directly to $W_n$, which generally will contain information about the character’s name since the name appears earlier in the sequence. This allows the model to backtrace effectively from $W_q$ through $W_n$ to the character. In the inverted template, $W_q$ may still attend to $W_n$, but since the fact is incomplete (the character hasn’t been mentioned yet), the model cannot use that attention to resolve the question. Instead, it must rely on weaker signals encoded in the character’s name to establish the link, which becomes harder with longer contexts due to diminishing attention strength. While these findings shed light on the challenge, deeper mechanistic analysis is beyond the scope of this paper and requires further study.
\addtocounter{footnote}{-1}
\begin{figure*}[t]
    \centering
    \subfigure[Full Sweep (One-hop)]{
        \includegraphics[width=0.23\textwidth, trim=10 10 10 8, clip] {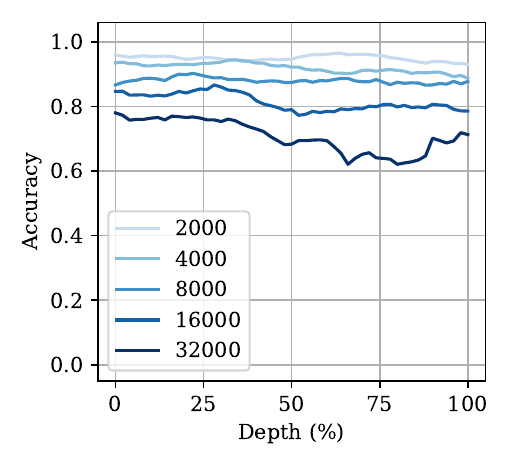}
        \label{fig:depth_full_onehop}
    }
    \subfigure[Full Sweep (Two-hop)]{
        \includegraphics[width=0.23\textwidth, trim=10 10 10 8, clip] {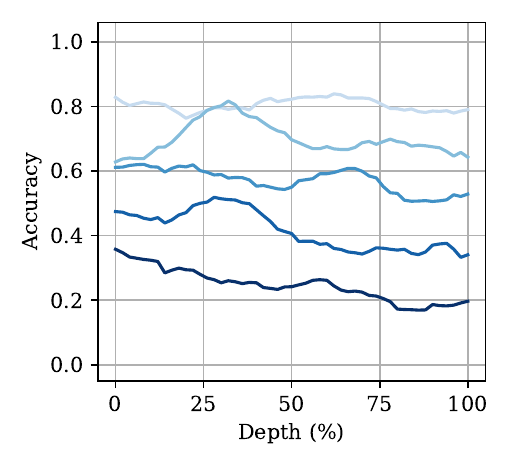}
        \label{fig:depth_full_twohop}
    }
    \subfigure[Last 2K (One-hop)]{
        \includegraphics[width=0.23\textwidth, trim=10 10 10 8, clip] {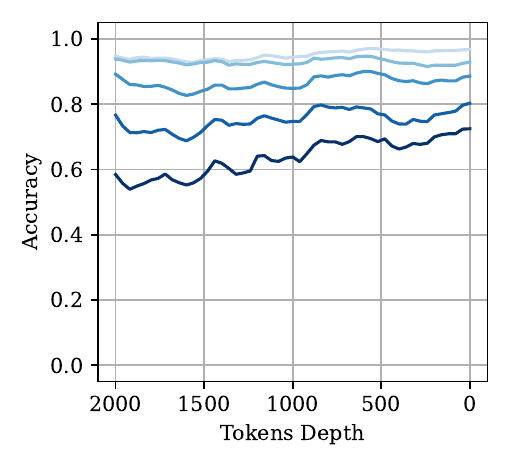}
        \label{fig:last_2k_onehop}
    }
    \subfigure[Last 2K (Two-hop)]{
        \includegraphics[width=0.23\textwidth, trim=10 10 10 8, clip] {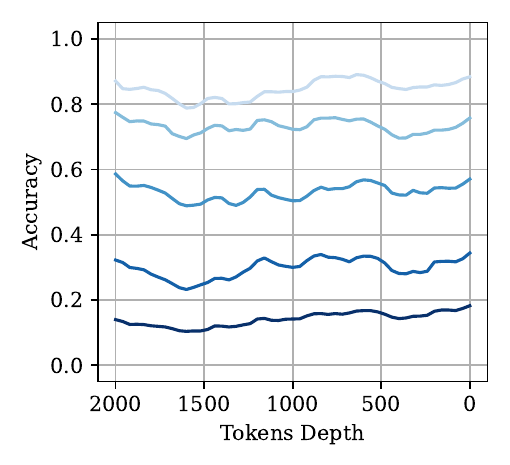}
        \label{fig:last_2k_twohop}
    }
    \caption{The full sweep plots (a \& b) illustrate performance across the entire context window, where 0\% corresponds to the beginning of the haystack and 100\% to the end. 
    The plots for the last 2K tokens (c \& d) depict performance when needle placements are aligned within that range for various context lengths; 0 marks the end of the context, and larger values indicate positions farther from the end (up to 2K tokens inward).
    The color shading of each plot line represents the tested context length. 
    To minimize noise and highlight trends more clearly, we increased the number of placements from 26 to 51 and applied a moving average with a window size of 12.\footnotemark}
    \label{fig:depth_plots}
\end{figure*}


\begin{figure}[b!]
    \centering
    \subfigure{
        \includegraphics[width=0.85\linewidth, trim=3 3 3 1, clip] {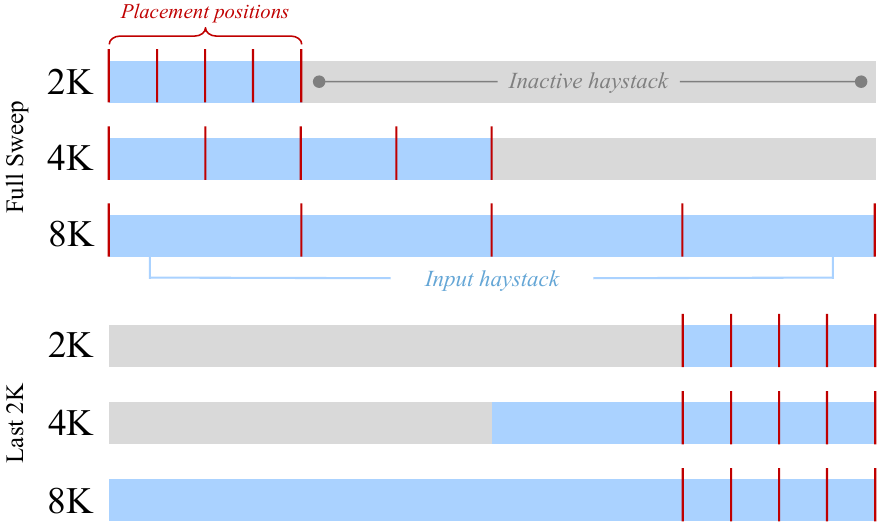}
    }
    \caption{Needle placements in full sweep (top) vs. last 2K tokens sweep (bottom): In the last 2K setup, placement positions are aligned in different context lengths, unlike the proportion-based positioning in full sweep.}
    \label{fig:full_vs_last2K_demonstrate}
\end{figure}
\subsubsection{Needle Placement Depth Analysis}
\label{sec:depth_analysis}
A common evaluation across NIAH-based benchmarks \cite{kamradt2023needle} examines the impact of needle placement within the context window. In Figure \ref{fig:depth_full_onehop}, we observe a ``lost-in-the-middle'' effect \cite{liu-etal-2024-lost} in 32K, where model performance dips when the needle appears in the middle of longer contexts.

Additionally, Figure \ref{fig:depth_full_twohop} reveals a key phenomenon: longer contexts in more complex (two-hop) examples dampen the performance distribution over the full sweep depending on their length. In vanilla multi-document or NIAH-based benchmarks \cite{kamradt2023needle, liu-etal-2024-lost}, models perform consistently well when the needle (or gold document) appears at the very beginning or end of the context window, with minimal impact from context length. 
However, in \framework, as task complexity increases in two-hop scenarios, larger context sizes shift the entire trendline downward toward zero, with performance declining even at the edges of the context window.

To further investigate this issue, we devise an alternative setup that focuses on analyzing the last 2K tokens instead of sweeping across the full context. Therefore, we align the placement positions in the last 2K tokens for all context lengths (see Figure \ref{fig:full_vs_last2K_demonstrate}).
This ensures that for a certain token depth the only
changing factor in each plotline is the context length,
which in turn means that the model has more tokens that it
needs to attend to.

Based on the final 2K results in Figure \ref{fig:last_2k_onehop}, the one-hop setup confirms our earlier observations from the full-sweep plots. The “lost-in-the-middle” phenomenon---where performance dips toward the center of the context---primarily appears in simpler tasks. Each plotline drops as it moves toward the center, reflecting its dependence on placement position and the way the model encodes positional information.
In contrast, the two-hop scenario appears to be influenced more by attention limitations than by position encoding alone. Figure \ref{fig:last_2k_twohop} reveals that, rather than depth exacerbating performance drops, the plot lines remain relatively stable over the last 2K positions. However, context length significantly reduces the overall performance trends observed in this range.
Llama 3.x models, like many other recent language models, feature rotary position embeddings (RoPE), a relative PE \cite{su2024roformer}. 
For each token depth in Figure \ref{fig:last_2k_twohop}, as the relative distance between question and fact remains the same regardless of context length, position encoding does not explain the performance drop.
Instead, the main limiting factor is the increased context length: as the number of tokens grows, the attention mechanism struggles to process information effectively. In the absence of strong surface-level cues (e.g., literal matches), locating relevant facts becomes challenging for the model, regardless of their position within long contexts.
\footnotetext{All figures use Llama 3.3 70B. Plots without smoothing applied are available in Appendix~\ref{sec:appx_depth_plots_raw}.}

\subsubsection{CoT Prompting}
\label{sec:cot_prompting}
Since \framework examples require an associative reasoning between the needle and question keywords to retrieve the correct answer, in this part we evaluate when the model is prompted to reason in a Chain-of-Thought (CoT) style \citep{CoT-Wei-2022} before returning a final answer (see Appendix \ref{sec:appx_task_prompts} for more details). 
In Table \ref{tab:results_w_reasoning}, we present the results when asked for CoT compared to asking directly for the final answer.
CoT prompting shows improvements over long-context tests and it shows a higher rate of improvement in two-shot. Despite the improvements, the tasks seem to remain challenging. For example, two-hop examples with CoT prompting barely achieve the scores of one-hop examples without CoT and continue to perform poorly on texts 16K tokens or longer. The challenge with CoT prompting is that the questions in \framework are straightforward. They are mentioning a singular clue to the answer, meaning they cannot be further decomposed into simpler steps. This limits the benefits of CoT prompting. However, the difficulty lies in reasoning through the association between the question and the needle, which remains a significant challenge for the model.

\begin{table}[t!]
        \small
	\setlength\tabcolsep{3pt}
	\centering
	\begin{tabular}{lcccc}
		\toprule
         & {\textbf{4K}} & {\textbf{8K}} & {\textbf{16K}} & {\textbf{32K}} \\
        \midrule
        &\multicolumn{4}{c}{\emph{One-hop}} \\
        - w/o CoT     & 90.3 & 84.1          & 73.2           & 56.2 \\
        - w/ CoT    & 95.6 & 91.1 & 82.6 & 60.6 \\
        Increase rate     & 5.9\% & 8.3\% & 12.8\% & 7.8\% \\
        \midrule
        &\multicolumn{4}{c}{\emph{Two-hop}} \\
        - w/o CoT    & 70.7 & 57.4 & 42.7 &  25.9 \\
        - w/ CoT    & 82.4 & 70.1 & 56.7 & 34.3 \\
        Increase rate     & 16.5\% & 22.1\% & 32.7\% & 32.4\% \\
		\bottomrule
	\end{tabular}
	\caption{Comparison of Chain-of-Thought (CoT) improvements in performance for Llama 3.3 70B, evaluated on both one-hop and two-hop tests.}
	\label{tab:results_w_reasoning}
\end{table}
To assess the performance of reasoning-based models (e.g., GPT-o1) on \framework, we selected the 10 most challenging needle-question pairs from the 58 available, based on the results summarized in Table \ref{tab:models_performance}.
We refer to this subset as \framework-Hard and present the evaluation results in Table \ref{tab:results_w_reasoningmodels}.
While reasoning-based models outperform CoT prompting on Llama 3.3, they still fail to achieve full-length generalization on this subset. Across all models, performance drops below the 50\% mark at 32K context length.
Notably, base scores are nearly perfect, demonstrating the simplicity of the task—even within this designated ``hard'' subset. 
This means that even with intermediate reasoning steps, models still struggle to link the needle to the question in long contexts without surface-level cues.

\subsubsection{Ablation Study: Literal match effect}
\label{sec:literal_match_effect}
To examine the simplifying impact of literal matches on results, we define two new sets of tests: (1) \textbf{Direct}: questions that explicitly ask about the fact stated in the needle by stating $W_n$ in the question, resembling a vanilla NIAH evaluation \cite{kamradt2023needle}. (2) \textbf{Multiple Choice (MC)}: questions that maintain the required latent associative reasoning while incorporating literal matches. In this setup, the question includes four character names as answer options—three from the haystack and one correct answer from the needle.

\begin{table}[t!]
        \small
	\setlength\tabcolsep{4pt}
	\centering
	\begin{tabular}{lccccc}
		\toprule
         & \textbf{Base} & \multirow{2}{*}{\textbf{4K}} & \multirow{2}{*}{\textbf{8K}} & \multirow{2}{*}{\textbf{16K}} & \multirow{2}{*}{\textbf{32K}} \\
         & \textbf{Score} &&&& \\
        \midrule
        &\multicolumn{5}{c}{\emph{Llama 3.3 70b}} \\
        - w/o CoT     & 98.3 & 55.5 & \cellcolor[HTML]{faa7a7}37.2 & \cellcolor[HTML]{faa7a7}16.7 & \cellcolor[HTML]{faa7a7}8.9 \\
        - w/ CoT    & 97.1 & 73.0 & 51.2 & \cellcolor[HTML]{faa7a7}31.8 & \cellcolor[HTML]{faa7a7}10.1 \\
        \midrule
        &\multicolumn{5}{c}{\emph{Reasoning models}} \\
        GPT-o1    & 99.9 & 92.0 & 78.0 & 60.1 & \cellcolor[HTML]{faa7a7}31.1 \\
        GPT-o3 Mini   &      98.8 & 52.8 & \cellcolor[HTML]{faa7a7}36.9 & \cellcolor[HTML]{faa7a7}25.5 & \cellcolor[HTML]{faa7a7}18.9 \\
        DeepSeek R1-DL-70b & 99.9 & 91.4 & 75.5 & \cellcolor[HTML]{faa7a7}49.4 & \cellcolor[HTML]{faa7a7}20.7 \\
	\bottomrule
	\end{tabular}
	\caption{Evaluation results of \framework-Hard: Scores falling below 50\% of the base score are highlighted in \colorbox{shadedRed}{red}.}
	\label{tab:results_w_reasoningmodels}
\end{table}
As expected, Table \ref{tab:literal_match_ablation} shows that direct examples with a high degree of literal overlap between the question and the needle are straightforward for the model to answer, even in long contexts, consistent with prior findings in RULER \cite{hsieh2024ruler}. Additionally, literal matches significantly aid the model when the questions remain unchanged, and only the multiple-choice format is introduced. The inclusion of literal matches in the multiple-choice setup provides significant guidance to the model. By offering the character names as answer options, including the correct name from the needle, the model can focus its search within a smaller scope. This dramatically simplifies the task of identifying the correct answer, as the literal match serves as a direct hint, reducing ambiguity in the reasoning process.
\begin{table}[b!]
        \small
	\setlength\tabcolsep{5pt}
	\centering
	\begin{tabular}{lccc}
		\toprule
          & {\textbf{8K}} & {\textbf{16K}} & {\textbf{32K}} \\
        \midrule
        Direct  & 98.3 & 98.5 & 98.5 \\
        \midrule
        One-hop & 84.1 & 73.2 & 56.2 \\
        \hspace{3pt}- w/ Literal Match (MC) & 98.7 & 97.4 & 93.1 \\
        \midrule
        Two-hop & 57.4 & 42.7 & 25.9 \\
        \hspace{3pt}- w/ Literal Match (MC) & 96.3 & 94.6 & 87.2 \\
		\bottomrule
	\end{tabular}
	\caption{Results in two literal match setups: direct and multiple choice (MC) questions. Model: Llama 3.3 70B}
	\label{tab:literal_match_ablation}
\end{table}

\begin{figure}[t!]
    \centering
    \subfigure{
        \includegraphics[width=0.6\linewidth, trim=10 8 10 8, clip] {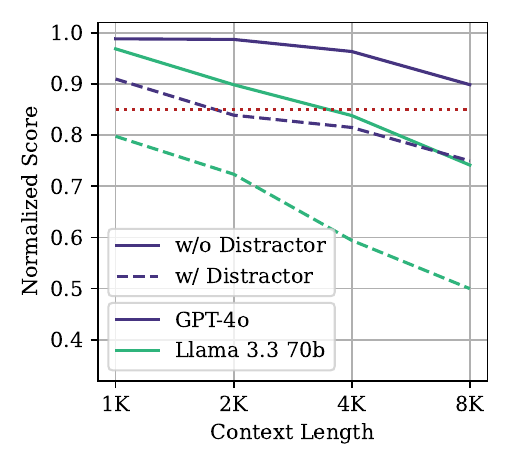}
    }
    \caption{Normalized performance comparison across GPT-4o and Llama 3.3 70B models, with and without distractors. The red dotted line marks the 0.85 effective threshold.}
    \label{fig:distractors}
\end{figure}

\paragraph{Distracting Literal Matches.} While literal matches  serve as cues if they are part of the relevant fact, they can also act as distractors if they are irrelevant to the answer. In Section \ref{sec:related_literal_matching}, we noted that some related benchmarks include similar documents in the context as distractors to test the model’s ability to discern the correct answer from irrelevant ones. This setup creates matches between the query and both relevant and irrelevant documents or facts. In contrast, \framework allows us to explore a different scenario: when the context contains distracting words overlapping with the question, while the relevant fact has minimal overlap with the query.
We insert a distractor sentence into the haystack (details in Appendix \ref{sec:appx_distractors}) containing $W_q$ but entirely irrelevant to both the needle and the question’s intent. This setup poses a significant challenge, requiring the model to disregard irrelevant literal overlaps while identifying a relevant fact with no meaningful overlap with the query.
As shown in Figure \ref{fig:distractors}, such distractors have a substantial impact on degrading length generalization. GPT-4o now demonstrates an effective length of just 1K, while Llama 3.3 70B performs even worse. While adding distractors slightly lowers base scores (GPT-4o: 93.8, Llama 3.3 70B: 84.4), the normalized plots still clearly illustrate a performance drop at longer lengths. These results highlight the challenge of resolving queries in contexts where irrelevant overlaps mislead the model, and the relevant fact shares no overlap with the question.

\section{Conclusion}
\framework provides a challenging benchmark for evaluating the reasoning capabilities of large language models in long-context settings. By removing literal overlaps between questions and relevant information, the benchmark tests models' ability to infer and link information within extensive irrelevant content. Our findings show that even state-of-the-art models struggle, especially as context length increases, revealing serious limits in their attention mechanism. While causal attention should theoretically access all previous tokens, models often rely on surface-level cues in longer contexts. This vulnerability becomes more pronounced when the context contains literal matches that fail to connect with the truly relevant fact, causing models to overlook the correct information and focus instead on superficial signals. 
We believe our findings with \framework are likely to extend to downstream applications. For instance, in search engines or RAG systems, a relevant document containing the correct answer may have a lexical gap with the query.
So, even if such a document is retrieved alongside others that likely have higher lexical similarity, language models may struggle to extract the correct answer, as they can become distracted by the lexical overlap with these other documents.
This work highlights the need for benchmarks that go beyond surface-level retrieval to assess deeper reasoning. 
\framework sets a new standard for evaluating long-context comprehension and emphasizes the importance of developing approaches capable of handling complex reasoning in long contexts.

\section*{Impact Statement}
This paper presents work aimed at advancing the field of long-context language modeling by evaluating and analyzing the most commonly used LLMs. There are many potential societal consequences of our work, none which we feel must be specifically highlighted here.

\section*{Acknowledgments}
We thank Abdullatif Köksal, Leonie Weissweiler, and Amir Hossein Kargaran for their valuable feedback and support, particularly in the early stages of this project. We also appreciate the insights from our peers, and
we are grateful to the anonymous reviewers for
their constructive comments.
\bibliography{paper_models, paper_related, paper_other}
\bibliographystyle{icml2025}

\newpage
\appendix
\onecolumn
\section{Needle Set Design \& Considerations}
\label{sec:appx_needle_set}
In Table \ref{tab:needle_set_full}, we demonstrate the full needle set that we use in \framework. In designing the needle templates, there are multiple considerations involved.
First, all templates in the needle set begin with a small introductory phrase or at least one word (e.g., ``Actually,'' ``In 2013,'') to distinguish themselves from the preceding context. This ensures that the needle's keyword or character is not inadvertently linked to the prior context. Since a newline is appended at the end of each needle, this issue is mitigated if the keyword or character appears at the end of the needle.

\begin{table*}[h!]
	\small
	\setlength\tabcolsep{5pt}
	\centering
    \begin{tabular}{p{0.25\linewidth} p{0.05\linewidth} p{0.37\linewidth} p{0.02\linewidth} p{0.18\linewidth}}
        \toprule
        \textbf{Question} & \multicolumn{2}{l}{\textbf{Needles}} & \multicolumn{2}{l}{\textbf{Keyword Types}}\\
        \midrule
        \multirow{4}{*}{Which character has been to $W_q$?} & Def. & There \,\,was\,\, [CHAR]\,\, who \,\,was \,\,an \,\,engineer\,\, living in $W_n$. & $W_n$ & Countries, cities, states\\
         & Inv. & There was an engineer living in $W_n$, named [CHAR]. & $W_q$ & Countries, cities, states \\
        \midrule
        \multirow{2}{*}{Which character has been to $W_q$?} & Def. & Actually, [CHAR] lives next to the $W_n$. & $W_n$ & Buildings \& Landmarks\\
         & Inv. & $W_n$ is next to where [CHAR] lives. & $W_q$ & Countries, cities, states \\
         \midrule
         \multirow{4}{*}{Which character has been to $W_q$?} & Def. & In 2013, after waiting in line for hours, [CHAR] finally saw the original $W_n$ painting up close. & $W_n$ & Buildings \& Landmarks\\
         & Inv. & In 2013, the original $W_n$ painting was seen up close by [CHAR], finally, after waiting in line for hours. & $W_q$ & Countries, cities, states \\
         \midrule
         \multirow{4}{*}{Which character cannot drink $W_q$?} & Def. & A message came in from [CHAR] saying, ``I'm $W_n$'' and nothing more. & $W_n$ & Dietary restriction\ \ \ \ \ \
         (e.g., lactose intolerant)\\
         & Inv. & A message came in saying, ``I'm $W_n$,'' from [CHAR]. & $W_q$ & Drinks \& Beverages \\
         \midrule
         \multirow{3}{*}{Which character cannot eat $W_q$?} & Def. & Then [CHAR] mentioned that he has been $W_n$ for years. & $W_n$ & Dietary restriction\ \ \ \ \ \  (e.g., vegan)\\
         & Inv. & There was a $W_n$ guest, named [CHAR]. & $W_q$ & Foods \\
        \bottomrule
    \end{tabular}
	\caption{Our proposed needle set templates in \framework. The placeholders [CHAR], $W_q$, and $W_n$ represent the randomly selected character (also the answer), the query keyword, and the needle keyword, respectively. Def.:
      default order. Inv.: inverted order.}
	\label{tab:needle_set_full}
\end{table*}

Another consideration is that the needle keyword should be uniquely associated with the query keyword. For instance, in the following sentence:
\begin{center}
There was an engineer living in \emph{Cambridge}, named Yuki.
\end{center}

Although the term "Cambridge" is commonly associated with the "United Kingdom," it is not uniquely so; it could also refer to cities in the United States, Canada, or other countries.
Additionally, we aim to avoid relying on language-specific markers. Many cities have distinctive elements in their names, such as orthographic features, morphological structures, or cultural naming conventions, that hint at their linguistic or geographic origins. By minimizing the influence of such markers, the needle design ensures a more rigorous evaluation of the model’s ability to make meaningful connections based on learned knowledge rather than surface-level linguistic cues.
For each template, we manually curated 2-6 keyword pairs, resulting in a total of 28 keyword pairs. Taking into account the order of fact statements, this generates 58 needle-question pairs.

\section{Models}
\label{sec:appx_models}
In Table \ref{tab:model_details}, we list all the models selected for evaluation. Models that are open weights were deployed using the vLLM library \cite{kwon2023efficient}, with weights obtained from HuggingFace \cite{wolf-etal-2020-transformers}.

\begin{table*}[h]
        \small
	\setlength\tabcolsep{5pt}
	\centering
	\begin{tabular}{lccl}
		\toprule
        \textbf{Model}  &   Context Length  & Open Weights? &   Model Revision \\
        \midrule
        GPT-4.1             & 1M   & No & \texttt{gpt-4.1-2025-04-14} \\ 
        GPT-4.1 mini        & 1M   & No & \texttt{gpt-4.1-mini-2025-04-14} \\ 
        GPT-4.1 nano        & 1M   & No & \texttt{gpt-4.1-nano-2025-04-14} \\ 
        \midrule
        GPT-4o              & 128K & No & \texttt{gpt-4o-2024-11-20} \\
        GPT-4o mini         & 128K & No & \texttt{gpt-4o-mini-20240718} \\
        \midrule
        Llama 4 Maverick    & 1M & Yes & \texttt{meta-llama/Llama-4-Maverick-17B-128E-Instruct} \\
        Llama 4 Scout       & 10M & Yes & \texttt{meta-llama/Llama-4-Scout-17B-16E-Instruct} \\
        \midrule
        Llama 3.3 70B       & 128K & Yes & \texttt{meta-llama/Llama-3.3-70B-Instruct} \\
        Llama 3.1 405B      & 128K & Yes & \texttt{meta-llama/Llama-3.1-405B-Instruct} \\
        Llama 3.1 70B       & 128K & Yes & \texttt{meta-llama/Llama-3.1-70B-Instruct} \\
        Llama 3.1 8B        & 128K & Yes & \texttt{meta-llama/Llama-3.1-8B-Instruct} \\
        \midrule
        Gemini 1.5 Pro      & 2M   & No     & \texttt{gemini-1.5-pro-002} \\
        Gemini 1.5 Flash    & 1M   & No     & \texttt{gemini-1.5-flash-002} \\ 
        Gemini 2.0 Flash    & 1M   & No     & \texttt{gemini-2.0-flash} \\
        Gemini 2.5 Flash    & 1M   & No     & \texttt{gemini-2.5-flash-preview-05-20} \\
        \midrule
        Gemma 3 27B       & 128K & Yes & \texttt{google/gemma-3-27b-it} \\
        Gemma 3 12B       & 128K & Yes & \texttt{google/gemma-3-12b-it} \\
        Gemma 3 4B        & 128K & Yes & \texttt{google/gemma-3-4b-it} \\     
        \midrule
        Claude 3.5 Sonnet   & 200K & No     & \texttt{anthropic.claude-3-5-sonnet-20241022-v2} \\ 
        \midrule
        Jamba 1.5 Mini      & 256K & Yes    & \texttt{ai21labs/AI21-Jamba-1.5-Mini} \\ 
        Command R+          & 128K & Yes    & \texttt{CohereForAI/c4ai-command-r-plus-08-2024} \\ 
        Mistral Large 2     & 128K & Yes    & \texttt{mistralai/Mistral-Large-Instruct-2411} \\
        \midrule
        \multicolumn{4}{c}{\emph{Reasoning-based models}} \\
        GPT-o1              & 128K & No    & \texttt{gpt-o1-2024-12-17} \\ 
        GPT-o3 Mini         & 128K & No    & \texttt{gpt-o3-mini-2025-01-31} \\ 
        DeepSeek R1-DL-70b  & 128K & Yes    & \texttt{deepseek-ai/DeepSeek-R1-Distill-Llama-70B} \\
	\bottomrule
	\end{tabular}
	\caption{Details of the selected models used for evaluation.}
	\label{tab:model_details}
\end{table*}

\section{Task Prompt Templates \& Inference Settings}
\label{sec:appx_task_prompts}
In Table \ref{tab:prompt_templates}, we present the task prompts used across all evaluations. While we do not employ the commonly used "Let's think step by step" prompt in the Chain-of-Thought (CoT) setup \cite{kojima2022large}, our prompt encourages the model to elaborate and expand its reasoning sufficiently before producing a final answer. To manage the extensive testing scope—7,540 tests per context length—we limit reasoning to three sentences or a maximum of 192 generated tokens.
In the CoT setup, a test is considered successful if the final answer (on the newline) includes the correct answer. This differs with the non-CoT setup, where success is determined based on whether the correct answer is present within the generated output.
For all standard instruction-tuned models, we use greedy decoding during generation. For reasoning-based models, we utilize the default sampling decoding mechanism for GPT-o1 and GPT-o3 Mini, while R1-based models employ top-P sampling with p = 0.95 and a temperature of 0.6. In addition, we cap the maximum number of generated tokens in reasoning-based models at 1536 tokens, including both reasoning and output tokens.
In all models, we apply each model's instruction-tuned chat templates.
\begin{table*}[h]
    \small
    \centering
    \begin{tabular}{>{\centering\arraybackslash}m{3cm}>{\RaggedRight\arraybackslash}p{10cm}}
    \toprule
    \textbf{Mode} & \textbf{Prompt Template} \\ \midrule
    \multirow{10}{*}{w/o CoT} & You will answer a question based on the following book snippet: \newline\vspace{0.3cm} \{haystack w/ needle\}\newline\vspace{0.3cm} Use the information provided in the book snippet to answer the question. Your answer should be short and based on either explicitly stated facts or strong, logical inferences.\newline\vspace{0.3cm} Question: \{question\}\newline\vspace{0.3cm} Return only the final answer with no additional explanation or reasoning. \\ \midrule
    \multirow{10}{*}{w/ CoT} & You will answer a question based on the following book snippet:\newline\vspace{0.3cm} \{haystack w/ needle\} \newline\vspace{0.3cm} Use the information provided in the book snippet to answer the question. Be aware that some details may not be stated directly, and you may need to INFER the answer based on the given information. Begin with a brief explanation of your reasoning in NO MORE THAN THREE (3) sentences. Then, return the final answer on a new line.\newline\vspace{0.3cm} Question: \{question\} \\
    \bottomrule
    \end{tabular}
    \caption{Details of prompt templates utilized in our evaluation.}
    \label{tab:prompt_templates}
\end{table*}

\section{Distractor Design}
\label{sec:appx_distractors}
To construct and integrate the distractor sentences mentioned in Section \ref{sec:literal_match_effect}, we devised two templates, applied uniformly across all needle-question pairs. Depending on the $W_q$, we use one of the following templates:
\begin{center} 
There was an article about $W_q$ in the daily newspaper.

or

There was a photo of $W_q$ in the daily newspaper.
\end{center}
Some instances of $W_q$ may naturally include an article (e.g., "a" or "an"), making them better suited for the second template, while others fit the first. Regardless of the choice, the templates are designed to remain neutral and unrelated to the intent of the question or the fact stated by any needle.

To minimize interference with the needle, we randomly place the distractor sentence while ensuring a token distance of at least 20\% of the context length. For example, in a 1K-token test, the distractor must be at least 200 tokens away from the needle.
Additionally, to avoid any advantage from proximity to the beginning or end of the context (which may gain extra attention), we restrict placement to between the 20\% and 80\% marks of the context length.
Together, these two constraints leave a span of 40\%-60\% of the context length available for random placement of the distractor sentence.
\begin{table*}[h!]
        \small
	\setlength\tabcolsep{5pt}
	\centering
	\begin{tabular}{l|cc|ccccccccc}
		\toprule
        \multirow{2}{*}{\textbf{Models}} & \textbf{Claimed} & \textbf{Effective} & \textbf{Base Score} & \multirow{2}{*}{\textbf{1K}} & \multirow{2}{*}{\textbf{2K}} & \multirow{2}{*}{\textbf{4K}} & \multirow{2}{*}{\textbf{8K}} & \multirow{2}{*}{\textbf{16K}} & \multirow{2}{*}{\textbf{32K}} & \multirow{2}{*}{\textbf{64K}} & \multirow{2}{*}{\textbf{128K}\footnotemark} \\
        & \textbf{Length} & \textbf{Length} & ($\times$0.85: \textbf{Thr.}) &&&&&&&& \\
        \midrule
        GPT-4.1            & 1M   & 16K & 97.0 (82.5) & \underline{95.6} & \underline{95.2} & \underline{91.7} & \underline{87.5} & \underline{84.9} & 79.8 & 69.7 & 64.7 \\
        GPT-4o             & 128K & 8K & 99.3 (84.4) & \underline{98.1} & \underline{98.0} & \underline{95.7} & \underline{89.2} & 81.6 & 69.7 & 62.4 & 56.0 \\
        Gemini 2.5 Flash   & 1M   & 2K & 94.4 (80.2) & \underline{90.1} & \underline{86.1} & 79.4 & 68.2 & 57.9 & 48.4 & --- & --- \\
        Gemini 2.0 Flash   & 1M   & 4K & 89.4 (76.0) & \underline{87.7} & \underline{87.5} & \underline{77.9} & 64.7 & 48.2 & \cellcolor[HTML]{faa7a7}41.0 & \cellcolor[HTML]{faa7a7}33.0 & \cellcolor[HTML]{faa7a7}16.4 \\
        Llama 4 Maverick   & 1M  & 2K & 90.1 (76.6) & \underline{81.6} & \underline{78.3} & 68.8 & 49.0 & \cellcolor[HTML]{faa7a7}34.3 & \cellcolor[HTML]{faa7a7}24.5 & \cellcolor[HTML]{faa7a7}--- & \cellcolor[HTML]{faa7a7}--- \\
        Gemma 3 27B        & 128K   & $<$1K & 88.6 (75.3) & 73.3 & 65.6 & 48.1 & \cellcolor[HTML]{faa7a7}32.7 & \cellcolor[HTML]{faa7a7}20.2 & \cellcolor[HTML]{faa7a7}9.5 & \cellcolor[HTML]{faa7a7}--- & \cellcolor[HTML]{faa7a7}--- \\
        Gemma 3 12B        & 128K   & 1K & 87.4 (74.3) & 74.7 & 61.8 & \cellcolor[HTML]{faa7a7}39.9 & \cellcolor[HTML]{faa7a7}27.4 & \cellcolor[HTML]{faa7a7}16.8 & \cellcolor[HTML]{faa7a7}7.3 & \cellcolor[HTML]{faa7a7}--- & \cellcolor[HTML]{faa7a7}--- \\
        Llama 4 Scout      & 10M  & 1K & 81.7 (69.4) & \underline{72.3} & 61.8 & 50.8 & \cellcolor[HTML]{faa7a7}35.5 & \cellcolor[HTML]{faa7a7}26.9 & \cellcolor[HTML]{faa7a7}21.6 & \cellcolor[HTML]{faa7a7}--- & \cellcolor[HTML]{faa7a7}--- \\
        GPT-4.1 Mini       & 1M   & $<$1K & 80.9 (68.8) & 66.7 & 62.8 & 58.7 & 51.9 & 46.2 & \cellcolor[HTML]{faa7a7}38.8 & \cellcolor[HTML]{faa7a7}--- & \cellcolor[HTML]{faa7a7}--- \\
        GPT-4.1 Nano       & 1M   & $<$1K & 80.7 (68.6) & 60.8 & 48.2 & \cellcolor[HTML]{faa7a7}36.7 & \cellcolor[HTML]{faa7a7}28.8 & \cellcolor[HTML]{faa7a7}19.5 & \cellcolor[HTML]{faa7a7}9.4 & \cellcolor[HTML]{faa7a7}--- & \cellcolor[HTML]{faa7a7}--- \\
        Gemma 3 4B         & 128K   & $<$1K & 73.6 (62.6) & 50.3 & \cellcolor[HTML]{faa7a7}35.3 & \cellcolor[HTML]{faa7a7}16.4 & \cellcolor[HTML]{faa7a7}7.5 & \cellcolor[HTML]{faa7a7}2.3 & \cellcolor[HTML]{faa7a7}0.9 & \cellcolor[HTML]{faa7a7}--- & \cellcolor[HTML]{faa7a7}--- \\
		\bottomrule
	\end{tabular}
	\caption{\framework benchmark results for GPT-4o, Gemini 2.0 Flash, and additional recent models. For models with stronger performance at 32K, we extend evaluation to 64K and 128K using the same setup. Scores above the effective threshold are underlined; scores below 50\% of the base score are shaded in \colorbox{shadedRed}{red}.}
	\label{tab:models_performance_w_64K_and_128K}
\end{table*}

\section{Results Beyond 32K \& Recent LLMs}
\label{sec:appx_beyond_32K}
Based on the dataset configuration outlined in Section \ref{sec:dataset_config}, we construct haystacks by randomly concatenating snippets extracted from the filtered books dataset. This setup enables evaluation across scalable context lengths, including 64K and 128K tokens, and \textbf{can be extended further} as model limits allow.

Two practical adjustments were made in this evaluation setup, without altering the core methodology: (1) Reduced Placement Count: Due to cost limitations, we reduce the number of needle placements from the default 26 to 11 placements per context length. This change minimizes API usage while still preserving meaningful coverage across the haystack.
(2) Token Limit Constraints: GPT-4o has a strict 128,000 token limit, including both input and output tokens. To accommodate the task prompt and model response, we limit the haystack length to 127,500 tokens for this model. To ensure a fair comparison, we use the same haystack length for Gemini 2.0 Flash.

\footnotetext{127,500-token haystack used based on model token limit constraints.}
The results of this evaluation are presented in Table \ref{tab:models_performance_w_64K_and_128K}, focusing on GPT-4o and Gemini 2.0 Flash at 64K and 128K context lengths. While GPT-4o does not fully meet its claimed performance at the maximum context length, it nonetheless maintains over 50\% of its base score at 128K, indicating relatively strong performance at that scale. In contrast, Gemini 2.0 Flash shows a sharper decline, dropping to 16.4\% at 128K.

To cover recent model releases, Table \ref{tab:models_performance_w_64K_and_128K} also includes entries for newer models such as the GPT-4.1 series and Gemini 2.5 Flash\footnote{Without reasoning (thinking budget = 0).} \citep{gemma_2025, team2024gemini}. For those that showed stronger performance at 32K, we also extend the evaluation to 128K using the same setup. GPT-4.1 shows clear improvements over prior models; however, its effective context length remains around 16K--well below the claimed 1M--and its performance drops below 65\% on 128K context lengths.

\section{One- \& Two-hop Results}
\label{sec:appx_one_and_two_hop_performance}
Tables \ref{tab:models_performance_one_hop} and \ref{tab:models_performance_two_hop} present detailed results on the one-hop and two-hop subsets of the \framework benchmark, evaluated across the selected models. The tables follow the same format and thresholding criteria as Table~\ref{tab:models_performance}, reporting both the claimed and effective context lengths. Note that the base scores--and consequently the thresholds--are computed separately for each subset. Although both one-hop and two-hop subsets show strong base scores, two-hop tasks generally yield shorter effective context lengths. This suggests that while models can perform well on complex reasoning in short contexts, their performance degrades more rapidly as context length increases, indicating reduced length generalization under greater reasoning demands.

\begin{table*}[ht!]
        \small
	\setlength\tabcolsep{5pt}
	\centering
	\begin{tabular}{l|cc|ccccccc}
		\toprule
        \multirow{2}{*}{\textbf{Models}} & \textbf{Claimed} & \textbf{Effective} & \textbf{Base Score} & \multirow{2}{*}{\textbf{1K}} & \multirow{2}{*}{\textbf{2K}} & \multirow{2}{*}{\textbf{4K}} & \multirow{2}{*}{\textbf{8K}} & \multirow{2}{*}{\textbf{16K}} & \multirow{2}{*}{\textbf{32K}} \\
        & \textbf{Length} & \textbf{Length} & ($\times$0.85: \textbf{Thr.}) &&&&&& \\
        \midrule
        GPT-4o              & 128K & 16K & 99.3 (84.4) & \underline{97.7} & \underline{97.5} & \underline{95.6} & \underline{91.9} & \underline{87.3} & 79.8 \\
        Llama 3.3 70B       & 128K & 8K & 97.7 (83.1) & \underline{97.1} & \underline{93.6} & \underline{90.3} & \underline{84.1} & 73.2 & 56.2 \\
        Llama 3.1 405B      & 128K & 4K & 91.7 (78.0) & \underline{88.7} & \underline{87.3} & \underline{80.2} & 68.4 & 59.6 & 49.4 \\
        Llama 3.1 70B       & 128K & 4K & 96.4 (82.0) & \underline{95.2} & \underline{89.8} & \underline{82.7} & 76.9 & 66.3 & 56.9 \\
        Gemini 1.5 Pro      & 2M   & 4K & 90.8 (77.2) & \underline{85.7} & \underline{85.5} & \underline{81.9} & 72.3 & 63.4 & 55.1 \\
        Jamba 1.5 Mini      & 256K & 2K & 93.1 (79.1) & \underline{80.0} & \underline{80.2} & 77.9 & 71.5 & 62.9 & 56.0 \\
        Command R+          & 128K & 2K & 91.6 (77.8) & \underline{79.4} & \underline{78.4} & 75.6 & 52.2 & \cellcolor[HTML]{faa7a7}26.9 & \cellcolor[HTML]{faa7a7}10.4 \\
        Gemini 2.0 Flash    & 1M   & 4K & 92.5 (78.6) & \underline{91.5} & \underline{93.5} & \underline{89.6} & 78.4 & 61.8 & 52.8 \\
        Mistral Large 2     & 128K & 4K & 83.3 (70.8) & \underline{82.5} & \underline{86.1} & \underline{80.3} & 62.5 & 44.1 & \cellcolor[HTML]{faa7a7}27.5 \\
        Claude 3.5 Sonnet   & 200K & 8K & 90.5 (76.9) & \underline{89.9} & \underline{91.6} & \underline{89.5} & \underline{78.2} & 61.1 & \cellcolor[HTML]{faa7a7}45.2 \\
        Gemini 1.5 Flash    & 1M   & 1K & 85.7 (72.9) & \underline{76.4} & 71.8 & 63.6 & 57.0 & 48.7 & \cellcolor[HTML]{faa7a7}41.3 \\
        GPT-4o mini         & 128K & 1K & 88.4 (75.2) & \underline{81.0} & 73.6 & 57.6 & 45.4 & \cellcolor[HTML]{faa7a7}30.2 & \cellcolor[HTML]{faa7a7}20.0 \\
        Llama 3.1 8B        & 128K & 1K & 83.0 (70.5) & \underline{75.7} & 69.3 & 60.7 & 49.6 & \cellcolor[HTML]{faa7a7}35.7 & \cellcolor[HTML]{faa7a7}22.7 \\
		\bottomrule
	\end{tabular}
	\caption{
    \framework benchmark results on one-hop examples. Base scores and effective lengths are computed using only the one-hop subset.
    Scores above the effective threshold are underlined, while scores that are below 50\% of the base score are shaded in \colorbox{shadedRed}{red}.}
	\label{tab:models_performance_one_hop}
\end{table*}

\begin{table*}[ht!]
        \small
	\setlength\tabcolsep{5pt}
	\centering
	\begin{tabular}{l|cc|ccccccc}
		\toprule
        \multirow{2}{*}{\textbf{Models}} & \textbf{Claimed} & \textbf{Effective} & \textbf{Base Score} & \multirow{2}{*}{\textbf{1K}} & \multirow{2}{*}{\textbf{2K}} & \multirow{2}{*}{\textbf{4K}} & \multirow{2}{*}{\textbf{8K}} & \multirow{2}{*}{\textbf{16K}} & \multirow{2}{*}{\textbf{32K}} \\
        & \textbf{Length} & \textbf{Length} & ($\times$0.85: \textbf{Thr.}) &&&&&& \\
        \midrule
        GPT-4o              & 128K & 8K & 99.3 (84.4) & \underline{98.7} & \underline{98.7} & \underline{95.8} & \underline{85.9} & 74.6 & 57.4 \\
        Llama 3.3 70B       & 128K & 1K & 96.7 (82.2) & \underline{90.6} & 79.8 & 70.7 & 57.4 & \cellcolor[HTML]{faa7a7}42.7 & \cellcolor[HTML]{faa7a7}25.9 \\
        Llama 3.1 405B      & 128K & 2K & 95.3 (81.0) & \underline{89.4} & \underline{82.0} & 67.4 & 49.8 & \cellcolor[HTML]{faa7a7}34.6 & \cellcolor[HTML]{faa7a7}23.8 \\
        Llama 3.1 70B       & 128K & 1K & 92.0 (78.2) & \underline{85.9} & 72.0 & 57.0 & \cellcolor[HTML]{faa7a7}45.2 & \cellcolor[HTML]{faa7a7}33.8 & \cellcolor[HTML]{faa7a7}26.4 \\
        Gemini 1.5 Pro      & 2M   & 1K & 94.9 (80.7) & \underline{87.1} & 79.4 & 67.4 & 53.6 & \cellcolor[HTML]{faa7a7}45.7 & \cellcolor[HTML]{faa7a7}39.6 \\
        Jamba 1.5 Mini      & 256K & $<$1K & 91.7 (77.9) & 71.8 & 66.6 & 62.0 & 50.7 & \cellcolor[HTML]{faa7a7}40.0 & \cellcolor[HTML]{faa7a7}28.4 \\
        Command R+          & 128K & $<$1K & 90.0 (76.5) & 74.0 & 67.4 & 54.7 & \cellcolor[HTML]{faa7a7}23.8 & \cellcolor[HTML]{faa7a7}14.3 & \cellcolor[HTML]{faa7a7}3.8 \\
        Gemini 2.0 Flash    & 1M   & 2K & 85.7 (72.8) & \underline{83.1} & \underline{80.0} & 63.6 & 47.8 & \cellcolor[HTML]{faa7a7}31.4 & \cellcolor[HTML]{faa7a7}26.4 \\
        Mistral Large 2     & 128K & 2K & 93.6 (79.5) & \underline{90.4} & \underline{84.7} & 64.7 & \cellcolor[HTML]{faa7a7}37.8 & \cellcolor[HTML]{faa7a7}18.4 & \cellcolor[HTML]{faa7a7}7.9 \\
        Claude 3.5 Sonnet   & 200K & 2K & 84.0 (71.4) & \underline{79.8} & \underline{74.5} & 63.0 & \cellcolor[HTML]{faa7a7}41.5 & \cellcolor[HTML]{faa7a7}26.9 & \cellcolor[HTML]{faa7a7}10.9 \\
        Gemini 1.5 Flash    & 1M   & $<$1K & 83.5 (71.0) & 59.1 & 49.0 & \cellcolor[HTML]{faa7a7}35.6 & \cellcolor[HTML]{faa7a7}28.8 & \cellcolor[HTML]{faa7a7}19.2 & \cellcolor[HTML]{faa7a7}12.9 \\
        GPT-4o mini         & 128K & $<$1K & 80.4 (68.3) & 51.4 & \cellcolor[HTML]{faa7a7}39.3 & \cellcolor[HTML]{faa7a7}27.6 & \cellcolor[HTML]{faa7a7}16.9 & \cellcolor[HTML]{faa7a7}8.8 & \cellcolor[HTML]{faa7a7}5.9 \\
        Llama 3.1 8B        & 128K & $<$1K  & 68.9 (58.6) & 53.4 & 36.0 & \cellcolor[HTML]{faa7a7}23.6 & \cellcolor[HTML]{faa7a7}10.0 & \cellcolor[HTML]{faa7a7}6.5 & \cellcolor[HTML]{faa7a7}3.8 \\
		\bottomrule
	\end{tabular}
	\caption{\framework benchmark results on two-hop examples. Base scores and effective lengths are computed using only the two-hop subset.
    Scores above the effective threshold are underlined, while scores that are below 50\% of the base score are shaded in \colorbox{shadedRed}{red}.}
	\label{tab:models_performance_two_hop}
\end{table*}

\begin{figure*}[t]
    \centering
    \subfigure[Full Sweep (One-hop)]{
        \includegraphics[width=0.23\textwidth, trim=10 10 10 8, clip] {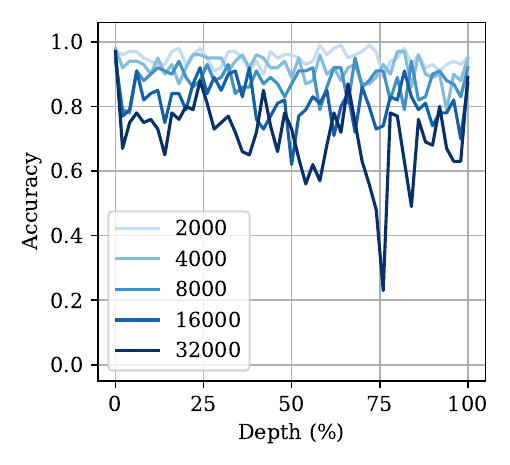}
        \label{fig:depth_full_onehop_raw}
    }
    \subfigure[Full Sweep (Two-hop)]{
        \includegraphics[width=0.23\textwidth, trim=10 10 10 8, clip] {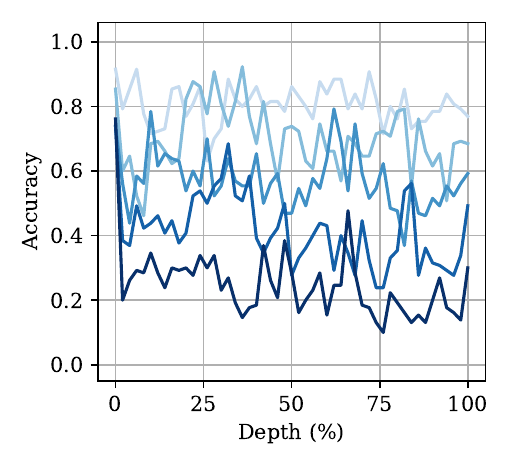}
        \label{fig:depth_full_twohop_raw}
    }
    \subfigure[Last 2K (One-hop)]{
        \includegraphics[width=0.23\textwidth, trim=10 10 10 8, clip] {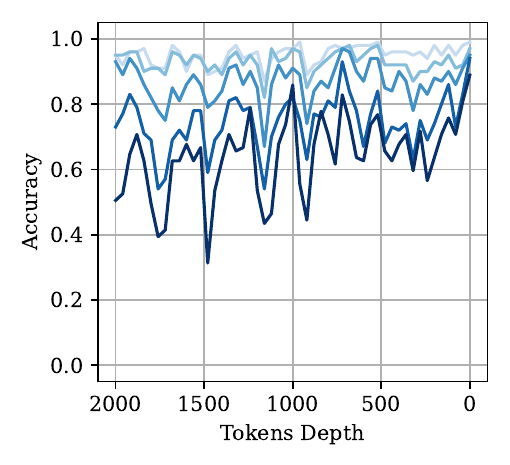}
        \label{fig:last_2k_onehop_raw}
    }
    \subfigure[Last 2K (Two-hop)]{
        \includegraphics[width=0.23\textwidth, trim=10 10 10 8, clip] {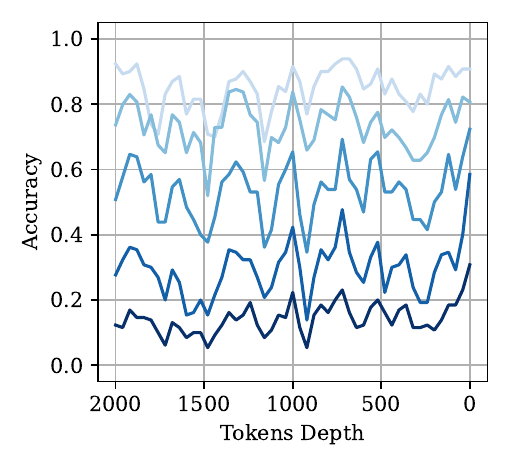}
        \label{fig:last_2k_twohop_raw}
    }
    \caption{Unsmoothed needle placement depth plots corresponding to the smoothed results in Figure \ref{fig:depth_plots}. These plots reflect raw performance values prior to applying the moving average.}
    \label{fig:depth_plots_raw}
\end{figure*}
\section{Raw Needle Placement Depth Plots}
\label{sec:appx_depth_plots_raw}
Figure \ref{fig:depth_plots_raw} presents the needle placement depth plots corresponding to Figure \ref{fig:depth_plots}, prior to the application of the moving average employed in the main figure.


\end{document}